\documentclass[10pt,twocolumn,letterpaper]{article}

\usepackage{cvpr}              % To produce the CAMERA-READY version
\definecolor{cvprblue}{rgb}{0.21,0.49,0.74}
\usepackage[pagebackref,breaklinks,colorlinks,allcolors=cvprblue]{hyperref}
\def\paperID{15} % *** Enter the Paper ID here
\def\confName{CVPR}
\def\confYear{2025}

\title{ReferGPT: Towards Zero-Shot Referring Multi-Object Tracking}

\author{
Tzoulio Chamiti\textsuperscript{1,2}\footnotemark[1]\quad
Leandro Di Bella\textsuperscript{1,2}\footnotemark[1] \quad
Adrian Munteanu\textsuperscript{1,2} \quad
Nikos Deligiannis\textsuperscript{1,2} \\
\textsuperscript{1}ETRO Department, Vrije Universiteit Brussel, Pleinlaan 2, B-1050 Brussels, Belgium \\
\textsuperscript{2}imec, Kapeldreef 75, B-3001 Leuven, Belgium \\
\tt\small{\textsuperscript{*}Equal Contributions}\\
{\tt\small \{tzoulio.chamiti, leandro.di.bella, adrian.munteanu, nikos.deligiannis\}@vub.be}\\
}

\begin{document}
\maketitle
\begin{abstract}
Tracking multiple objects based on textual queries is a challenging task that requires linking language understanding with object association across frames. Previous works typically train the whole process end-to-end or integrate an additional referring text module into a multi-object tracker, but they both require supervised training and potentially struggle with generalization to open-set queries. In this work, we introduce ReferGPT, a novel zero-shot referring multi-object tracking framework. We provide a multi-modal large language model (MLLM) with spatial knowledge enabling it to generate 3D-aware captions. This enhances its descriptive capabilities and supports a more flexible referring vocabulary without training. We also propose a robust query-matching strategy, leveraging CLIP-based semantic encoding and fuzzy matching to associate MLLM generated captions with user queries. Extensive experiments on Refer-KITTI, Refer-KITTIv2 and Refer-KITTI+ demonstrate that ReferGPT achieves competitive performance against trained methods, showcasing its robustness and zero-shot capabilities in autonomous driving. The codes are available on \url{https://github.com/Tzoulio/ReferGPT}
% The ABSTRACT is to be in fully justified italicized text, at the top of the left-hand column, below the author and affiliation information.
% Use the word ``Abstract'' as the title, in 12-point Times, boldface type, centered relative to the column, initially capitalized.
% The abstract is to be in 10-point, single-spaced type.
% Leave two blank lines after the Abstract, then begin the main text.
% Look at previous \confName abstracts to get a feel for style and length.
% \vspace{-0.5cm}
\end{abstract}    
\section{Introduction}
\label{sec:intro}
\begin{figure}[h]
  \centering
  \includegraphics[width=1.0\linewidth]{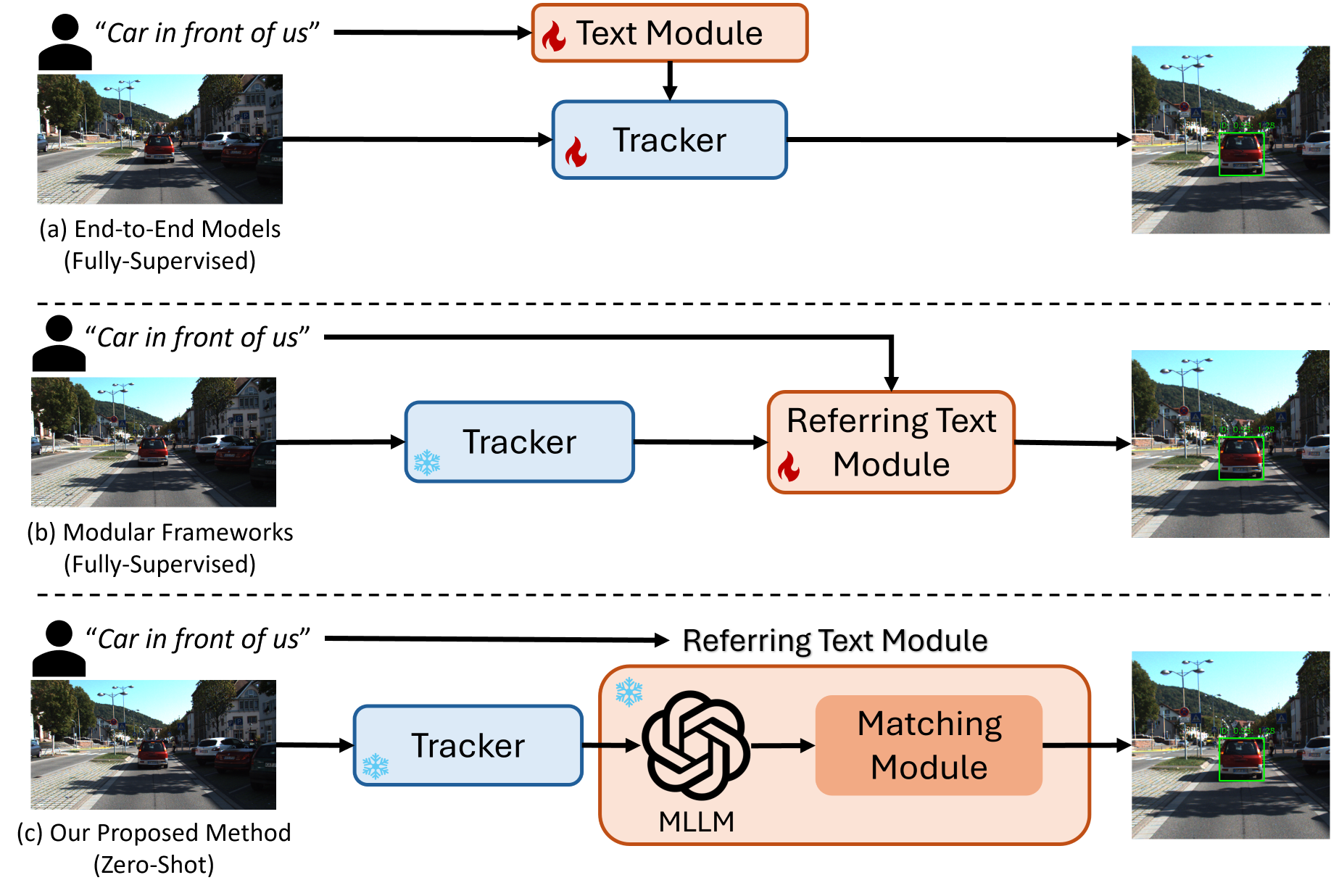}
   \caption{Comparison between ReferGPT and previous RMOT methods. (a) End-to-end models jointly learn tracking and referring. (b) Tracking-by-detection frameworks follow a modular approach and train the referring text module. (c) Our method builds on (b) while eliminating the need for training, enabling zero-shot referring MOT.}
   \label{fig:frameworks}
   \vspace{-0.3cm}
\end{figure}
Referring Multi-Object Tracking (RMOT) has emerged as a crucial problem in computer vision, particularly in autonomous driving, surveillance and human-machine interaction. Unlike the standard Multi-Object Tracking (MOT) task~\cite{geiger2012we, sun2020scalability}, which associates detections across frames without explicit user guidance, RMOT introduces textual queries to specify which objects to track \cite{wu2023referring, zhang2024bootstrapping, lin2024echotrack}. For instance, given a user query such as “the blue car turning right”, an RMOT system must identify and track the relevant object across a sequence of frames. This capability is particularly valuable in autonomous driving and intelligent traffic monitoring, where integrating natural language guidance enhances situational awareness and enables more intuitive human-machine interaction in complex environments.

% \begin{figure}[h]
%   \centering
%   \includegraphics[width=1.0\linewidth]{Figures/training_paradigm.png}
%    \caption{Comparison between ReferGPT and previous RMOT methods. (a) End-to-end models jointly learn tracking and referring but suffer from task interference. (b) Tracking-by-detection pipelines improve modularity but require retraining. (c) Our method builds on (b) while eliminating the need for retraining, enabling zero-shot referring MOT.}
%    \label{fig:onecol}
% \end{figure}

Generally, RMOT approaches can be divided into two categories, as seen in Fig.\ref{fig:frameworks}. The first category (Fig.\ref{fig:frameworks}(a)) consists of end-to-end transformer-based frameworks, such as TransRMOT~\cite{wu2023referring}, where a single supervised unified model performs both object tracking and referring. The second category, depicted in Fig.\ref{fig:frameworks}(b), includes modular frameworks that follow a tracking-by-detection paradigm, such as iKUN~\cite{du2024ikun}, where a dedicated supervised referring module is introduced into an existing tracker. However, while both categories have shown promising results, they remain constrained by two fundamental limitations. First, supervised RMOT methods tend to generalize poorly due to their reliance on a closed-set of predefined queries, limiting their ability to handle open-set ones. Second, 2D-based methods struggle with spatial alignment, particularly for queries that contain explicit spatial information. For example, queries such as “cars which are faster than ours” or “the car is moving away from us”, are difficult to refer to using only 2D representations. Additionally, end-to-end joint-detection-and-tracking methods often experience task interference, which degrades association performance. Tracking-by-detection methods remain more reliable, as separating detection and association allows for independent optimization of each task \cite{di2025hybridtrack, wang2024mctrack}. 

To address these limitations, we propose ReferGPT, a zero-shot referring MOT framework that combines multi-object tracking through Kalman filtering \cite{welch1995introduction} with a Multi-Modal Large Language Model (MLLM)-based referring modules for query matching. Unlike traditional RMOT methods \cite{wu2023referring, zhang2024bootstrapping, du2024ikun} that operate solely in the 2D space, our approach leverages spatial information from a 3D tracker \cite{wu20213d} to enrich the MLLM's understanding of the scene, allowing it to generate structured object captions, which are in turn matched with the appropriate user queries. Specifically, we incorporate 2D images as visual cues, together with 3D motion and position knowledge through the MLLM prompt, to generate semantically rich yet spatially aware object captions. This is critical for handling queries with depth-dependent constraints, such as "blue cars that are moving". 

Furthermore, to match the MLLM caption with the user query, we introduce a matching module that combines semantic understanding through CLIP encoding \cite{radford2021learningtransferablevisualmodels}  with deterministic data association using fuzzy matching. This allows us to align captions and queries even when they use semantically related terms, like "the vehicle in front of our car" with "the automobile ahead of us" providing a more robust matching process. In summary, our contributions are as follows. 
\begin{itemize}
    \item We introduce ReferGPT, a zero-shot Referring Multi-Object Tracking framework capable of handling arbitrary textual queries, without requiring training, for both 2D and 3D inputs. To the best of our knowledge, this is the first work on zero-shot RMOT in autonomous driving.  
    \item We demonstrate the effectiveness of Multi-Modal Large Language Models (MLLMs) in generating spatially grounded text, which enhances query-based tracking performance. 
    \item We perform extensive evaluations on the Refer-KITTI, Refer-KITTIv2 and Refer-KITTI+ datasets, demonstrating that our framework achieves competitive performance without relying on supervised training, highlighting its zero-shot capabilities on the autonomous driving setting.
    \item We conduct multiple ablation studies showing that each component of our proposed method contributes non-trivially. 
\end{itemize}

\section{Related Work}
\label{sec:formatting}
\textbf{Multi-Object Tracking}: Multi-Object Tracking methods can be broadly categorized into Joint Detection and Tracking (JDT) \cite{tokmakov2021learning}, \cite{tokmakov2022object}, \cite{wu2021tracklet}, \cite{10777493}, \cite{meinhardt2022trackformer}, \cite{zhou2020tracking} and Tracking-by-Detection (TBD) \cite{1211504}, \cite{weng20203d}, \cite{wu20213d}, \cite{cao2023observation}, \cite{wang2023camo} approaches. JDT methods combine detection and tracking into a unified framework, jointly optimizing object localization and association across frames. Some JDT methods operate in 2D, such as CenterTrack \cite{zhou2020tracking}, which integrates spatio-temporal memory for short-term association, and PermaTrack \cite{tokmakov2021learning}, which maintains object locations under full occlusions. Others extend JDT to 3D, such as MMF-JDT \cite{10777493}, which integrates object detection and multi-object tracking into a single model, eliminating the traditional data association step by predicting trajectory states and PC-TCNN \cite{wu2021tracklet} which generates tracklet proposals, refines them, and associates them to perform multi-object tracking. However, these approaches demonstrate performance limitations compared to TBD methods. TBD methods decouple detection from tracking, allowing for independent optimization and greater flexibility. Specifically, HybridTrack \cite{di2025hybridtrack} combines deep learning with a learnable Kalman Filter to dynamically adjust motion parameters, MCTrack \cite{yi2024ucmctrack} employs a two-stage matching process that combines bird’s-eye view and image-plane matching to improve robustness against depth errors and PC3T \cite{wu20213d} uses a confidence guided data association module for the tracking task. This motivates our choice of TBD as our tracking paradigm, which offers greater flexibility and improved performance.

\textbf{Referring Multi-Object Tracking}: The first work on Referring Multi-Object Tracking (RMOT), TransRMOT~\cite{wu2023referring}, introduces an end-to-end transformer framework that uses language expressions as semantic cues to track referred objects frame by frame. TempRMOT \cite{zhang2024bootstrapping} extends this by adding a temporal enhancement module to better handle motion across frames, while DeepRMOT~\cite{he2024visual} incorporates deep cross-modal fusion to improve how visual and linguistic features interact. ROMOT \cite{li2024romot} further leverages multi-stage cross-modal attention and vision-language modeling, enabling the tracking of both known and novel objects based solely on descriptive attributes. MLS-Track \cite{ma2024mls} enhances cross-modal learning by progressively integrating semantic information into visual features at multiple stages of the model. MGLT \cite{chen2025multi} combines linguistic, temporal, and tracking cues to generate object queries and enhance visual-language alignment. All of the above methods follow an end-to-end approach, integrating tracking and referring into a unified model to leverage cross-modal interactions, but they often lack flexibility. 

iKUN \cite{du2024ikun} was the first method to deviate from the traditional RMOT paradigm by introducing a modular approach. Specifically, iKUN proposed using an existing pre-trained tracker and combining it with a trainable referring text module. MEX \cite{tran2025mex} later extended this idea and further optimized the cross-modality attention for better computational efficiency of the framework. However, both approaches train the referring module, which limits their adaptability and requires retraining when faced with out-of-distribution queries. In contrast, ReferGPT operates in a zero-shot setting and removes the need to train the referring text module. This allows for more flexible referring, without the need for task-specific supervision or retraining.

\begin{figure*}[t]
  \centering
 
  \includegraphics[width=1.0\linewidth]{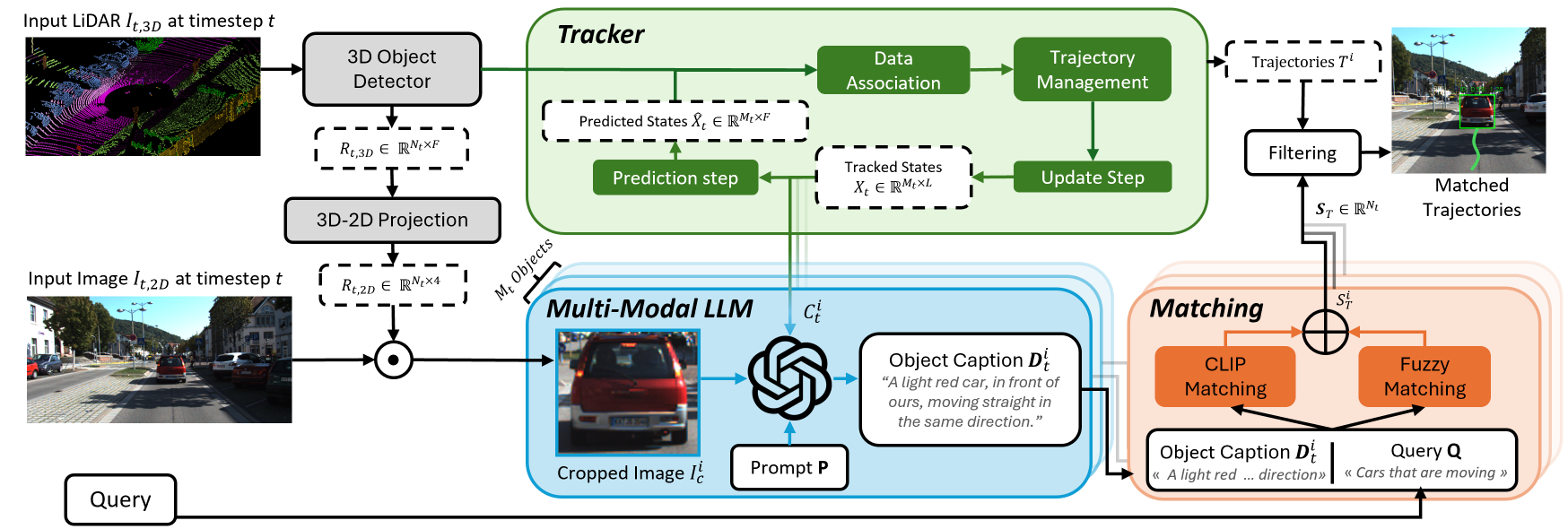}
   \caption{Overview of the proposed ReferGPT framework. Given LiDAR $I_{t,3D}$ and image $I_{t,2D}$ inputs, a 3D object detector extracts object candidates ${R}_{t,3D}$, which are then tracked using a tracking-by-detection approach with a Kalman filter for trajectory prediction. A Multi-Modal Large Language Model generates descriptive captions $\mathbf{D^i_t}$ for each object by leveraging object coordinates $C^i$  and appearance features $I_c^i$. These captions are then matched against the referring query \textbf{Q} using a matching module. The final matched trajectories $T^i$ are filtered and associated with the query to produce the final output. }
   \label{fig:main_arch}
\end{figure*}
\section{Methodology}

\subsection{Problem Formulation and Method Overview}
Given a sequence of $K$ frames \(\mathbf{I} = \{I_t\}_{t=1}^{K}\) and a referring text query $\mathbf{Q} = [q_1, q_2, \dots, q_m] \in \mathbb{R}^m$ where each $q_i$ represents a word in the sequence, our goal is to identify and track objects that match the given query. Using an off-the-shelf object detector $\mathcal{F_{\text{det}}(\cdot)}$, we generate a set of $N_t$ candidate detections per frame. Our method is designed to work with any object detector that produces 3D outputs, regardless of whether the input data comes from LiDAR or RGB images. Next, we employ a tracking-by-detection framework $\mathcal{F_{\text{trk}}(\cdot)}$, as shown in Fig.\ref{fig:main_arch}, which identifies and keeps track of all true detections. For every identified object, we use an MLLM agent to generate a textual caption $D$, leveraging 3D motion and position information from the tracker along with the cropped image $I_c$ of the detected object. The generated description and the referring query are then processed by our matching module, which computes a similarity score $S_T$ to determine their alignment. After all frames have been processed, we filter and associate the set of detected objects to the input query, through clustering, based on their computed similarity score.

In the following sections, we describe the usage of $\mathcal{F_{\text{det}}(\cdot)}$ and $\mathcal{F_{\text{trk}}(\cdot)}$ in Sec.\ref{sec:det_track}. Our MLLM agent is introduced in Sec.\ref{sec:mllm}. We detail our matching module in Sec.\ref{sec:matching} and we present our filtering process in Sec.\ref{sec:cluster}. 

\subsection{3D Object Detection and Multi-Object Tracking} \label{sec:det_track}
The tracking-by-detection approach tracks objects by associating detections extracted from a 3D object detector at each timestep \( t \). Formally, given an input frame \( I_t \) at timestep \( t \), the detector \( \mathcal{F_{\text{det}}}(\cdot) \) produces a set of \( N_t \) detections:
\begin{equation}
    {R}_t = \mathcal{F_{\text{det}}}(I_t), \quad {R}_t = \{\mathbf{r}^i_t\}_{i=1}^{N_t} \in \mathbb{R}^{N_t \times F}
\end{equation}
 where \( {R}_t \) represents the set of detected objects, \( N_t \) is the number of detections at timestep \( t \), and \( F \) is the number of attributes describing each detection. Each detected object \( \mathbf{r}^i_t \in \mathbb{R}^F \) corresponds to a 3D bounding box, parameterized as: $\mathbf{r}^i_t = [x, y, z, w, l, h, \theta] \in \mathbb{R}^7$ where \( (x, y, z) \) are the 3D centroid coordinates, \( (w, l, h) \) are the width, length, and height, and \( \theta \) is the heading angle. 

These detections serve as the basis for tracking objects over time. To maintain and update object trajectories, we employ a Kalman filter-based \cite{welch1995introduction} multi-object tracking module denoted as $\mathcal{F}_{\text{trk}}(\cdot)$. At each timestep $t$, $\mathcal{F}_{\text{trk}}(\cdot)$ initializes new trajectories for objects that have been newly detected and do not correspond to any existing track. Additionally, for objects that are already being tracked, it predicts their future states $\hat{X}_t = \{\mathbf{\hat{x}}^i_t\}_{i=1}^{M_t} \in \mathbb{R}^{M_t\times F}$ where $M_t$ is the number of objects currently being tracked at time $t$. Each trajectory $T^i$ for object $i$ is defined as a sequence of its estimated states $T^i = \left[\mathbf{x}^i_{t_{\text{init}}}, \dots, \mathbf{x}^i_t \right]$ where $t_{\text{init}}$ denotes the timestep when object $i$ was first detected and its trajectory was initialized. 

Next, detections ${R}_t$ are associated with $\hat{X}_t$ using a confidence-guided association strategy \cite{wu20213d}, which prioritizes high-confidence matches to improve robustness against false positives and occlusions. Once detections are assigned, tracked states, denoted as ${X}_t = \{\mathbf{{x}}^i_t\}_{i=1}^{M_t} \in \mathbb{R}^{M_t\times F}$, are updated by incorporating the new associated measurements, refining object localization. Finally, a trajectory management process handles track initiation, termination, and re-identification to ensure stable tracking, even in cases of temporary occlusion or missing detections.

\subsection{Multi-Modal Large Language Agent}\label{sec:mllm}
We define the Multi-Modal Large Language Agent as $\mathcal{F}_\text{LLM}(I^i_c, \mathbf{P}, \mathbf{C}^i_t)$ as shown in Fig.\ref{fig:main_arch}, where ${I^i_c}$ is the cropped image of the specific object and $\mathbf{P}$ denotes a predefined prompt provided to the MLLM in the form of a text sequence $\mathbf{P} = [p_1, p_2, \dots, p_n] \in \mathbb{R}^n$ with each $p_i$ being a word in the prompt. $\mathbf{C}^i_t$ is a compact representation of the $i$-th object's spatial and motion statistics over the most recent $T$ frames. To ensure spatial consistency, we perform a coordinate transformation that converts tracked object states  ${X}_t$ from the global (world) coordinate system to an ego-centric reference frame, aligned with the ego vehicle. Specifically, the tracked states $x_t^i \in$ ${X}_t$ are transformed into relative positions $\mathbf{p}^i_t \in \mathbb{R}^3$.

In this ego-centric coordinate system, $\mathbf{C}^i_t \in \mathbb{R}^9$ consists of the current position $\mathbf{p} = (x, y, z)$ at $t_0$, with $t_0$ denoting the current frame, the average heading angle $\bar{\theta}$ of the past T=5 frames, the Euclidean distance between the current state $\textbf{x}_{t_0}$ and the state at frame $t_0-T$, the mean heading angle variation $\bar{\Delta \theta}$, and the spatial variations $\Delta \mathbf{p} = (\Delta x, \Delta y, \Delta z)$ computed across the time window $T$: 
\begin{equation}
    \mathbf{C}^i_t = \left[ \mathbf{p}, \bar{\theta}, d_\text{euclid}, \Delta\mathbf{p}, \bar{\Delta \theta}\right],
\end{equation}
 This compact representation is then flattened into a 1-D text sequence and provided as input to the MLLM. By summarizing both the object’s current state and its recent motion over the temporal window $T$, $\mathbf{C}^i_t$ enables the agent to reason about dynamic behaviors such as being stationary, moving forward, parking, or turning. The agent, in turn, generates a descriptive output text sequence $\mathbf{D}^i_t = [d_1, d_2, \dots, d_n]$ $\in \mathbb{R}^n$ for each tracked object, where each $d_i$ represents a word in the sequence. %i is an exponent in \mathbf{D}^i_t

In summary, leveraging the LLM’s reasoning and natural language generation capabilities, we translate the spatial and kinematic properties of each object into a natural language description. These descriptions capture attributes such as object color, object type (e.g., car, pedestrian), relative location with respect to the camera, movement status, and directional information. The resulting descriptions serve as intermediate representations of the scene through text and are later used during the matching process, where they are compared with the referring query to identify the corresponding object.

\subsection{Matching}\label{sec:matching}
\begin{figure}[h]
  \centering
  \includegraphics[width=1.0\linewidth]{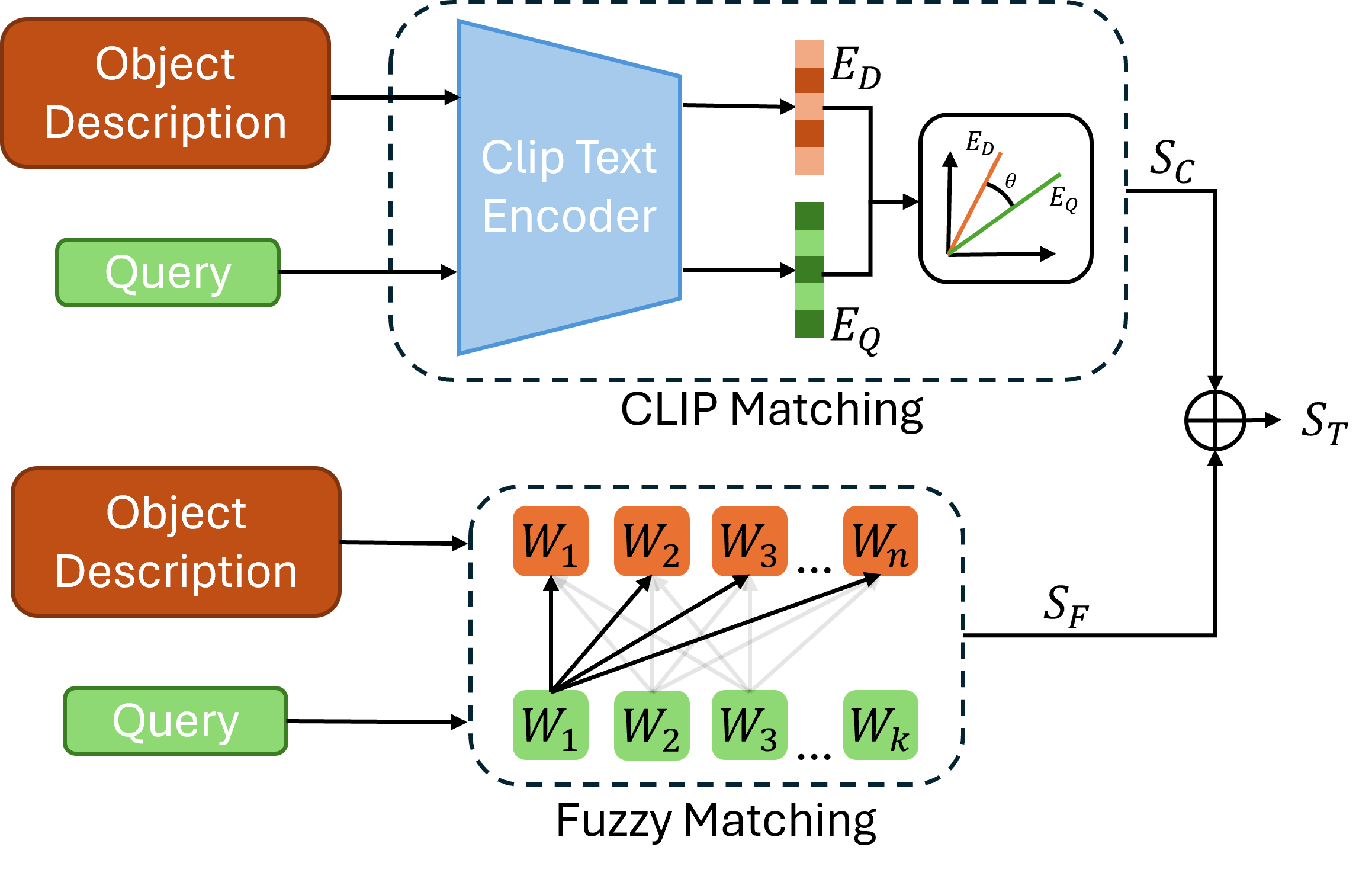}
   \caption{Our Matching Module. Given an object description $\textbf{D}_t^i $ and a referring query \textbf{Q}, we calculate the total matching score $S_T$ between them.}
   \label{fig:onecol}
% \vspace{-0.5cm}
\end{figure}

As shown in Fig.\ref{fig:onecol}, to compute the similarity between each detected object's description generated by the MLLM agent \( \mathbf{D}^i_t \) and the referring query \( \mathbf{Q} \), we employ a hybrid matching approach that combines fuzzy matching with semantic embedding-based similarity. We first apply the Ratcliff/Obershelp algorithm \cite{ratcliff1988pattern} to compute a fuzzy matching score between each word $q_k$ in the query $\mathbf{Q}$ and each word $d_k$ in the object description $\mathbf{D}^i_t$. The algorithm identifies the longest common contiguous subsequences of characters between word pairs to determine their similarity score. The similarity score for a pair of words $(q_k, d_k)$ is defined as: \begin{equation} S_k = \frac{2 \cdot M_k}{|q_k| + |d_{k}|}, \end{equation} where $M_k$ represents the total number of matching characters in the longest common subsequences, and $|q_k|$ and $|d_{k}|$ are the lengths of the query word and the description word, respectively. Finally, the overall fuzzy matching score $S_F$ between the query $\mathbf{Q}$ and the description $\mathbf{D}^i_t$ is defined as the sum of all per-word scores: $S_F = \sum_{k=1}^{m} S_k$

This technique prioritizes structural similarity by emphasizing shared contiguous substrings. However, fuzzy matching remains limited in handling synonyms and ambiguous terms, as it relies on direct character sequence comparisons rather than true semantic understanding. 
To address this limitation, we leverage the textual encoder of CLIP to encode both the object description and the referring query. Specifically, we obtain vector embeddings \( E_D \) and \( E_Q \) for \( \mathbf{D}^i_t \) and \( \mathbf{Q} \), respectively, and compute their similarity using the cosine similarity metric:
\begin{equation}
S_C = \frac{E_D \cdot E_Q}{\|E_D\| \|E_Q\|},
\end{equation}
where \( S_C \) represents the CLIP similarity score. By incorporating CLIP, we enhance the robustness of the matching process, enabling the model to handle synonyms, contextual variations, and ambiguous phrases that fuzzy matching alone may struggle to resolve. The final matching score is computed as $S_T = S_C + S_F$.

\subsection{Filtering Optimization}\label{sec:cluster}
Upon track termination, each associated detection $\mathbf{x}^i_t$ is characterized by its coordinates and a similarity score $S^i_T$ inferred during the matching step, indicating whether the detected object's description aligns with the given textual query $\textbf{Q}$. To mitigate the impact of noisy detections and reduce false negatives, we first apply a majority voting strategy: for every tracked object, if the majority of detections within the object trajectory $T_i$ are classified as matching the query, the entire trajectory is inferred to correspond to the query, effectively suppressing false negative matched detections. 

Following this filtering step, a key challenge arises from the logit-based similarity scores produced by the matching module. Since the similarity score varies across different queries, depending on query length and the response characteristics of the $\mathcal{F}_\text{LLM}$, using a fixed threshold for all queries is suboptimal. An alternative approach could be to select the top-$k$ tracks with the highest similarity scores. However, this strategy assumes prior knowledge of how many detections truly satisfy the query, which is unknown. To address this, we employ an agglomerative hierarchical clustering algorithm \cite{bouguettaya2015efficient} to dynamically identify the subset of tracks having the highest similarity scores. This adaptive clustering approach ensures that the most relevant trajectories are selected without relying on a rigid threshold, thereby enhancing the flexibility and generalization of the matching process.

\begin{table*}[h!]
\centering
\caption{Comparison of existing methods on Refer-KITTI dataset \cite{wu2023referring}. The best is marked in \textbf{bold}, and the second-best in \underline{underline}. 'E' indicates End-to-End methods. The results are reported in \%. $\mathbf{\text{ReferGPT}_\text{3D}}$ uses LiDAR as input, $\mathbf{\text{ReferGPT}_\text{2D}}$ uses Image as input.}
\label{tab:refer_v1}
\resizebox{0.99\textwidth}{!}{%
\begin{tabular}{lcc|ccc|ccc|c}
    \toprule
     & & & \multicolumn{3}{c|}{\textbf{Detection}} & \multicolumn{3}{c|}{\textbf{Association}} &  \\
    \textbf{Method} & \textbf{E} & \textbf{HOTA} $\uparrow$ & \textbf{DetA} $\uparrow$  & \textbf{DetRe}  $\uparrow$ & \textbf{DetPr} $\uparrow$ & \textbf{AssA} $\uparrow$ & \textbf{AssRe} $\uparrow$ & \textbf{AssPr} $\uparrow$ & \textbf{LocA}\\
    \midrule
    % \multicolumn{10}{l}{\textbf{Traditional Methods}}  \\
    \textbf{Traditional Methods}    &  &  &  &  &  &  &  &   &  \\
    FairMOT \cite{zhang2021fairmot}   & × & 22.78 & 14.43 & 16.44 & 45.48 & 39.11 & 43.05 & 71.65  & 74.77 \\
    DeepSORT  \cite{wojke2017simple}  & × & 25.59 & 19.76 & 26.38 & 36.93 & 34.31 & 39.55 & 61.05 & 71.34 \\
    ByteTrack  \cite{zhang2022bytetrack}  & × & 24.95 & 15.50 & 18.25 & 43.48 & 43.11 & 48.64 & 70.72 & 73.90 \\
    TransTrack  \cite{sun2020transtrack}  & × & 32.77 & 23.31 & 32.33 & 42.23 & 45.71 & 49.99 & 78.74 & 79.48 \\
    \midrule
    \textbf{Supervised Learning Methods}    &  &  &  &  &  &  &  &   &  \\

    iKUN \cite{du2024ikun}    & × & 48.84 & 35.74 & 51.97 & 52.26 & \textbf{66.80} & 72.95 & 87.09 & - \\
    MEX \cite{tran2025mex}    & × & 45.07 & 32.81 & \textbf{62.52} & 41.65 & - & 71.09 & - & - \\
    TransRMOT \cite{wu2023referring}    & \checkmark & 46.56 & 37.97 & 49.69 & \underline{60.10} & 57.33 & 60.02 & \textbf{89.67} & \underline{90.33} \\
    EchoTrack \cite{lin2024echotrack}   & \checkmark & 48.86 & \underline{41.26} & 53.42 & \textbf{62.83} & 57.59 & 61.61 & \underline{89.33} & \textbf{90.74} \\
    DeepRMOT \cite{he2024visual}   & \checkmark & 39.55 & 30.12 & 41.91 & 47.47 & 53.23 & 58.47 & 82.16 & 80.49 \\
    TempRMOT  \cite{zhang2024bootstrapping}   & \checkmark & \textbf{52.21} & \textbf{43.73} & \underline{55.65} & 59.25 & \underline{66.75} & {71.82} & 87.76 & 90.40 \\
    ROMOT  \cite{li2024romot}   & \checkmark & 35.5 & 28.30 & - & - & 46.2  & - & - & - \\
    MGLT  \cite{chen2025multi}   & \checkmark & 49.25 & 37.09 & - & - & 65.50  & - & - & - \\
    \midrule
    \textbf{Zero-Shot Learning Methods}    &  &  &  &  &  &  &  &   &  \\
    Baseline$^{*}$  &  × & {21.42} & {9.48} & {16.10} & {18.20} & {48.82} & {57.86} & {72.57} & 80.72 \\
    $\mathbf{\textbf{ReferGPT}_\textbf{2D}}$ (Ours)  &  × & {46.36} & {36.58} & {51.40} & {52.16} & {59.00} & \underline{73.16} & {69.31} & 83.26 \\
    $\mathbf{\textbf{ReferGPT}_\textbf{3D}}$ (Ours)  &  × & \underline{49.46} & {39.43} & {50.21} & {58.91} & {62.57} & \textbf{73.74} & {72.78} & 81.85\\
    \bottomrule
\end{tabular}}

%Why is DetRe decreasing when going from 2D to 3D?

\vspace{1mm}
\begin{minipage}{\textwidth}
\footnotesize
$^{*}$ \textit{Baseline} uses only CLIP-Image encoder for similarity evaluation, without the MLLM-Agent.
\end{minipage}
\end{table*}

\begin{table*}[h!]
\centering
 \caption{Comparison of existing methods on Refer-KITTIv2 dataset \cite{zhang2024bootstrapping}. The best is marked in \textbf{bold}, and the second-best in \underline{underline}. 'E' indicates End-to-End methods. The results are reported in \%.}
\label{tab:comparison_of_methods_refer_kittiv2}
\resizebox{0.99\textwidth}{!}{%
    \begin{tabular}{lcc|ccc|ccc|c}
    \toprule
     & & &  \multicolumn{3}{c|}{\textbf{Detection}} & \multicolumn{3}{c|}{\textbf{Association}} &  \\
    \textbf{Method} & \textbf{E} & \textbf{HOTA} $\uparrow$ & \textbf{DetA} $\uparrow$  & \textbf{DetRe}  $\uparrow$ & \textbf{DetPr} $\uparrow$ & \textbf{AssA} $\uparrow$ & \textbf{AssRe} $\uparrow$ & \textbf{AssPr} $\uparrow$ & \textbf{LocA}\\
    \midrule
    \textbf{Traditional Methods}    &  &  &  &  &  &  &  &   &  \\
FairMOT \cite{zhang2021fairmot}   & × & 22.53 & 15.80 & 20.60 & \underline{37.03} & 32.82 & 36.21 & 71.94  & 78.28 \\
ByteTrack \cite{zhang2022bytetrack}  & × & 24.59 & 16.78 & 22.60 & 36.18 & 36.63 & 41.00 & 69.63 & 78.00 \\ 
\midrule
    \textbf{Supervised Learning Methods}    &  &  &  &  &  &  &  &   &  \\
iKUN \cite{du2024ikun}    & × & 10.32 & 2.17 & 2.36 & 19.75 & 49.77 & 58.48 & 68.64 & 74.56 \\
TransRMOT \cite{wu2023referring}    & \checkmark & \underline{31.00} & \underline{19.40} & \textbf{36.41} & 28.97 & 49.68 & 54.59 & \textbf{82.29} & \underline{89.82} \\
TempRMOT \cite{zhang2024bootstrapping}    & \checkmark & \textbf{35.04} & \textbf{22.97} & \underline{34.23} & \textbf{40.41} & \underline{53.58} & \underline{59.50} & \underline{81.29} & \textbf{90.07} \\
\midrule
\textbf{Zero-Shot Learning Methods}    &  &  &  &  &  &  &  &   &  \\
\textbf{ReferGPT} (Ours) & × & {30.12} & {15.69} & {21.55} & {34.41} & \textbf{59.02} & \textbf{74.59} & {68.20} & {79.76}\\
\bottomrule
\end{tabular}}
\end{table*}

\begin{table*}[h!]
\centering
 \caption{Comparison of existing methods on Refer-KITTI+ dataset \cite{lin2024echotrack}. The best is marked in \textbf{bold}, and the second-best in \underline{underline}. 'E' indicates End-to-End methods. The results are reported in \%.}
\label{tab:comparison_of_methods_refer_kitti+}
\resizebox{0.99\textwidth}{!}{%
    \begin{tabular}{lcc|ccc|ccc|c}
    \toprule
     & & &  \multicolumn{3}{c|}{\textbf{Detection}} & \multicolumn{3}{c|}{\textbf{Association}} &  \\
    \textbf{Method} & \textbf{E} & \textbf{HOTA} $\uparrow$ & \textbf{DetA} $\uparrow$  & \textbf{DetRe}  $\uparrow$ & \textbf{DetPr} $\uparrow$ & \textbf{AssA} $\uparrow$ & \textbf{AssRe} $\uparrow$ & \textbf{AssPr} $\uparrow$ & \textbf{LocA}\\
    \midrule
    \textbf{Traditional Methods}    &  &  &  &  &  &  &  &   &  \\
TransRMOT \cite{wu2023referring}   & \checkmark & 35.32 & 25.61 &  40.05 & 38.45 & 50.33 & \underline{55.40} & \underline{81.23}  & 79.44 \\

EchoTrack \cite{lin2024echotrack}   & \checkmark & \underline{37.46} & \underline{28.83} & \textbf{39.83} & \underline{46.70} & \underline{50.39} & 54.14 & \textbf{82.57}  & \underline{79.97} \\
\midrule
\textbf{Zero-Shot Learning Methods}    &  &  &  &  &  &  &  &   &  \\

\textbf{ReferGPT} (Ours) & × & \textbf{43.44} & \textbf{29.89} & \underline{36.59} & \textbf{56.98} & \textbf{63.60} & \textbf{ 75.20 } & {73.27} & \textbf{82.23}\\
\bottomrule
\end{tabular}}
\end{table*}

\section{Experimental Setup}
\textbf{Dataset and Evaluation Metrics.}  We conduct our experiments on three benchmark datasets derived from Refer-KITTI. First, we evaluate on the Refer-KITTI-v1 public dataset \cite{wu2023referring}, using its test split, which consists of 3 videos and 150 diverse natural language queries. Second, we assess our method on Refer-KITTI-v2 \cite{zhang2024bootstrapping}, an extension of v1 featuring a more challenging set of queries. We report results on its pre-defined test split, comprising 4 videos and 859 queries, to demonstrate our method’s robustness in more complex settings. Additionally, we evaluate on Refer-KITTI+ \cite{lin2024echotrack}, following the same split protocol as EchoTrack \cite{lin2024echotrack}, which includes 3 videos and 154 queries. 

We primarily focus on the Higher Order Tracking Accuracy (HOTA) metric \cite{luiten2021hota}, which provides a balanced measure of detection, association and localization accuracy in multi-object tracking. It is defined as $\text{HOTA} = \sqrt{\text{DetA} \cdot \text{AssA}}$.
DetA is the Detection Accuracy \cite{luiten2021hota} emphasizing the accuracy of object detection, while AssA is the Association Accuracy \cite{luiten2021hota} explicitly measuring the association's effectiveness in maintaining track consistency. We also provide the Localization Accuracy (LoCA) which describes how accurately the objects' spatial positions are estimated.

\noindent \textbf{Implementation Details.} In our work, we leverage LiDAR input frames for 3D object detection and employ CasA as our object detector~\cite{wu2022casa}. We deploy the model-based PC3T tracker~\cite{wu20213d} within a tracking-by-detection framework to associate detections across frames. As our MLLM, we employ GPT-4o-mini for its enhanced reasoning and captioning capabilities. For textual encoding, we use CLIP ViT-L/14~\cite{radford2021learningtransferablevisualmodels}, which has demonstrated improved representation capabilities due to its transformer-based architecture and larger model size.

\section{Experiments}
Tab.~\ref{tab:refer_v1} presents a comprehensive comparison of existing methods on the Refer-KITTI dataset. Our proposed method demonstrates competitive performance despite operating in a zero-shot setting. Unlike other models that rely on task-specific training, ReferGPT generalizes to the referring multi-object tracking task without any training. Notably, by using 3D LiDAR data as input, we achieve a HOTA score of 49.46\%. When using 2D as input, our method scores 46.36\% HOTA. We also achieve the highest Association Recall (AssRe) scores, highlighting the ability to consistently maintain object identities over time. Despite being zero-shot, ReferGPT delivers competitive results even across detection and localization metrics, with strong DetA and LocA scores.

Tab.~\ref{tab:comparison_of_methods_refer_kittiv2}  and Tab.~\ref{tab:comparison_of_methods_refer_kitti+} present the performance of our proposed method on the Refer-KITTIv2 \cite{zhang2024bootstrapping} and Refer-KITTI+ \cite{lin2024echotrack} datasets, respectively. These datasets introduce additional, more complex queries, posing significant challenges for models not explicitly trained on them. Despite this, ReferGPT demonstrates strong generalization capabilities in handling such open-set referring queries. Specifically, we achieve a competitive  HOTA score of 30.12\%, closely matching supervised end-to-end approaches such as TransRMOT \cite{wu2023referring} 31.00\%. Notably, ReferGPT scores the highest association accuracy (AssA) of 59.02\%, significantly surpassing both TempRMOT \cite{zhang2024bootstrapping}  and TransRMOT \cite{wu2023referring}. For the Refer-KITTI+ dataset, ReferGPT sets a new benchmark in HOTA 43.44\%, in DetA 29.89\% and AssA 63.60\%, reaffirming our method's potential in open-set query tracking.

\begin{figure*}[h!]
    \centering
    % First row
    \begin{minipage}[b]{0.33\linewidth}
        \centering
        \includegraphics[width=\linewidth,trim={0cm 2.5cm 0cm 0.5cm},clip]{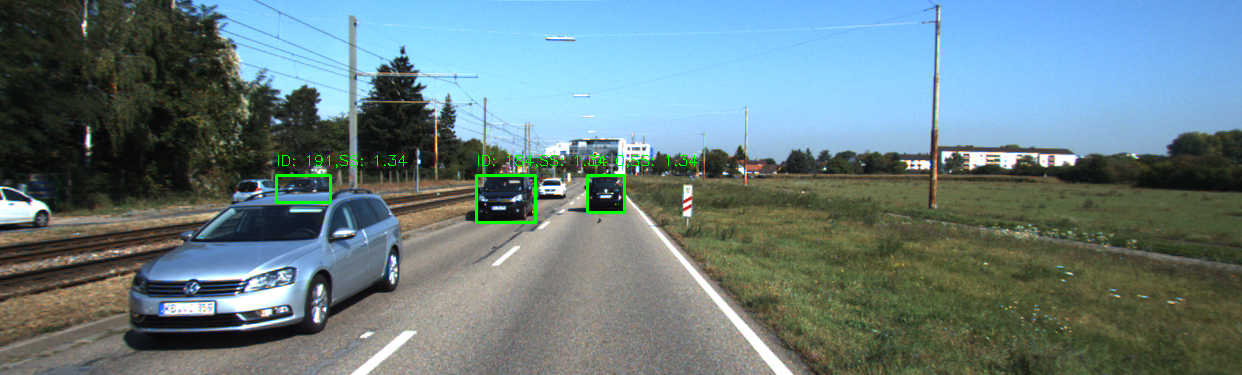}
    \end{minipage}%
    \begin{minipage}[b]{0.33\linewidth}
        \centering
        \includegraphics[width=\linewidth,trim={0cm 2.5cm 0cm 0.5cm},clip]{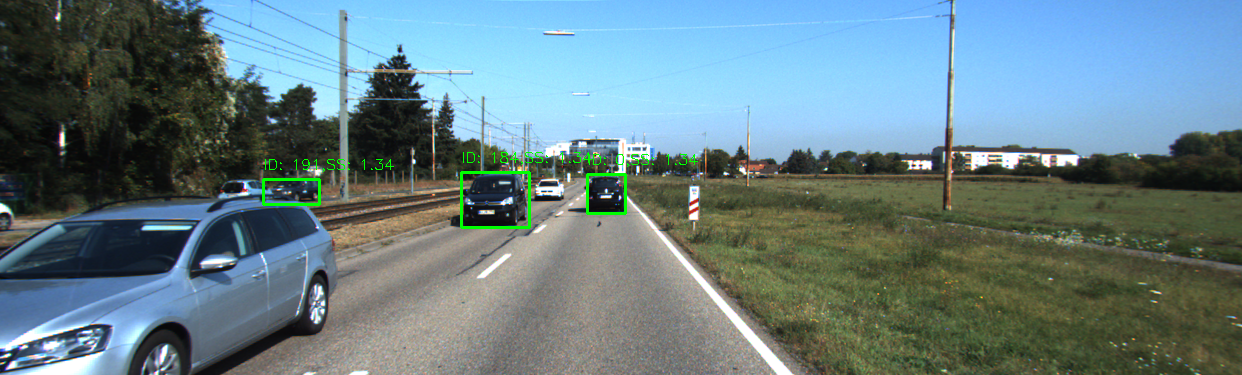}
    \end{minipage}%
    \begin{minipage}[b]{0.33\linewidth}
        \centering
        \includegraphics[width=\linewidth,trim={0cm 2.5cm 0cm 0.5cm},clip]{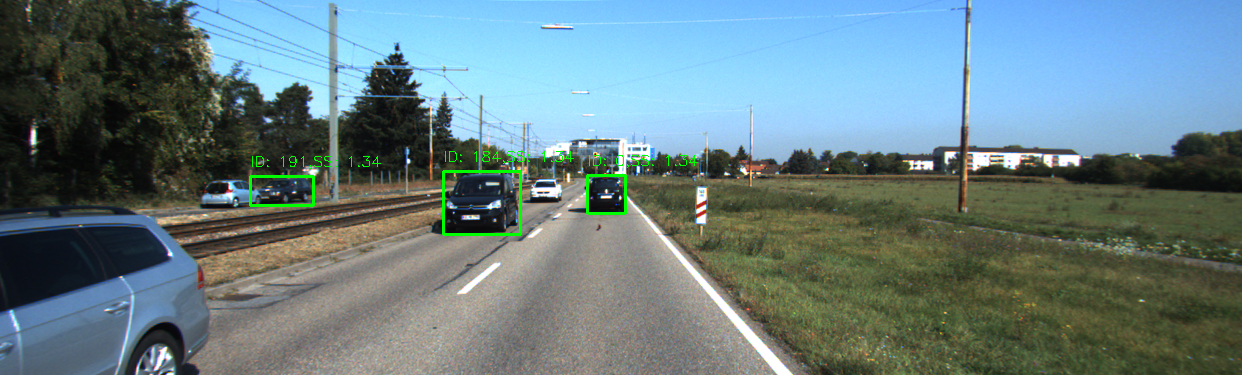}
    \end{minipage}\\
    \vspace{-0.3cm}
    
    \caption*{Query: Car in black}
    \vspace{0.25cm}

    % Second row
    \begin{minipage}[b]{0.33\linewidth}
        \centering
        \includegraphics[width=\linewidth,trim={0cm 2.5cm 0cm 0.5cm},clip]{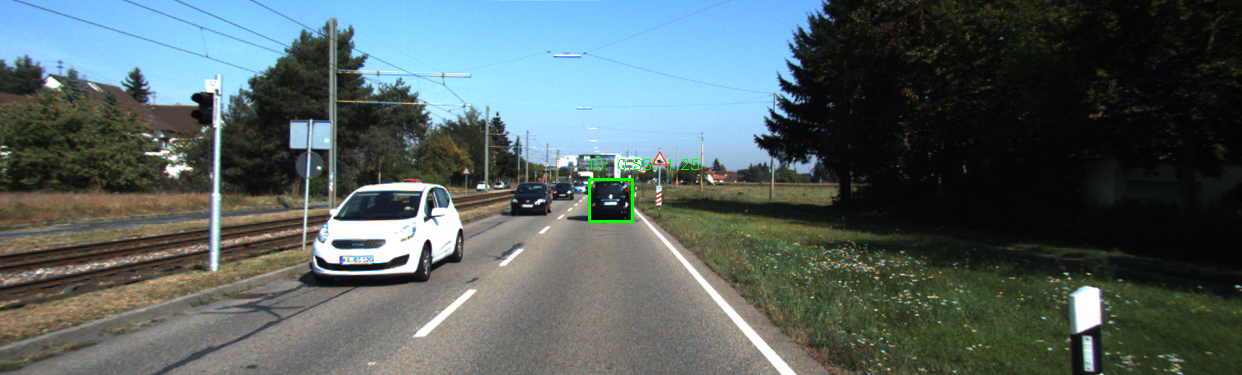}
    \end{minipage}%
    \begin{minipage}[b]{0.33\linewidth}
        \centering
        \includegraphics[width=\linewidth,trim={0cm 2.5cm 0cm 0.5cm},clip]{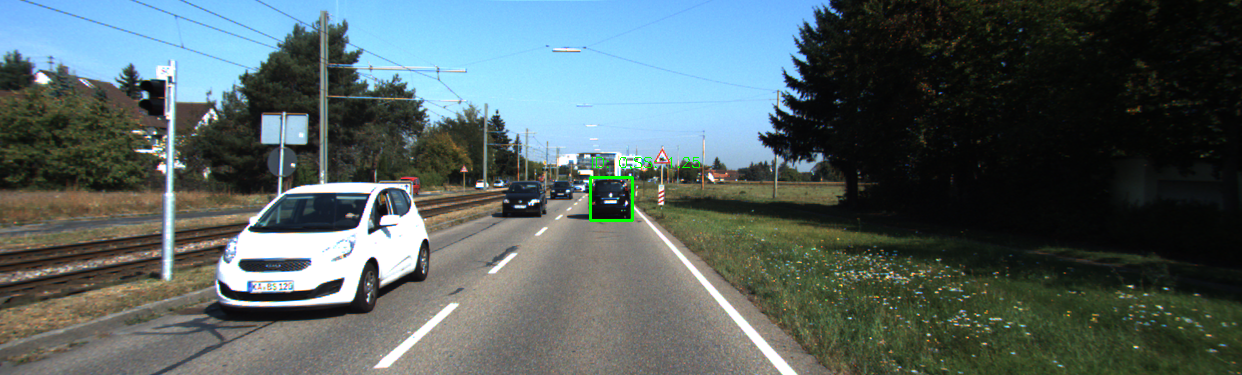}
    \end{minipage}%
    \begin{minipage}[b]{0.33\linewidth}
        \centering
        \includegraphics[width=\linewidth,trim={0cm 2.5cm 0cm 0.5cm},clip]{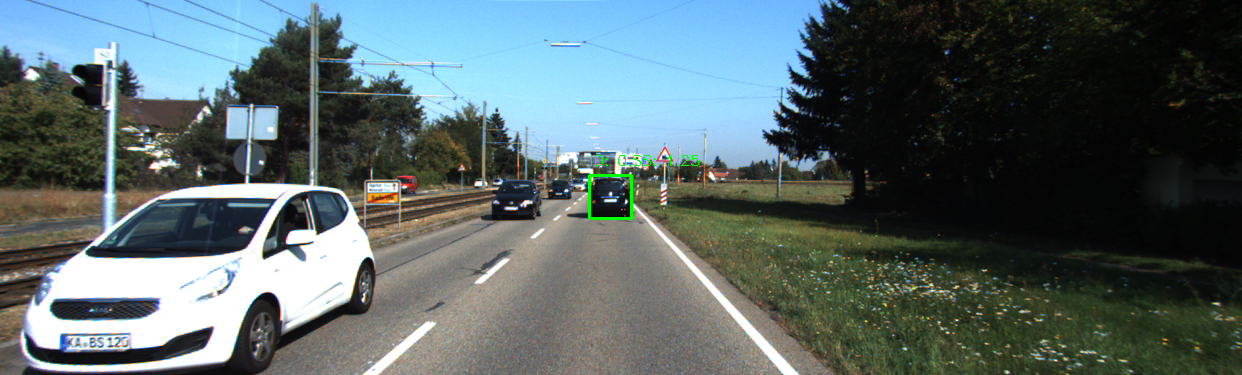}
    \end{minipage}\\
    \vspace{-0.23cm}
    
    \caption*{Query: Car in front}
    \vspace{0.25cm}

    % Third row
    \begin{minipage}[b]{0.33\linewidth}
        \centering
        \includegraphics[width=\linewidth,trim={0cm 0.3cm 0cm 1.0cm},clip]{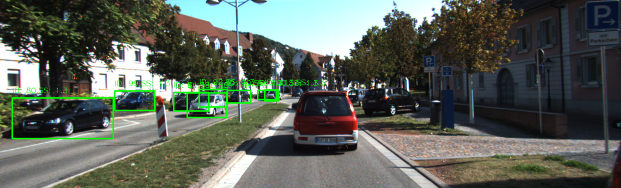}
    \end{minipage}%
    \begin{minipage}[b]{0.33\linewidth}
        \centering
        \includegraphics[width=\linewidth,trim={0cm 0.3cm 0cm 1.0cm},clip]{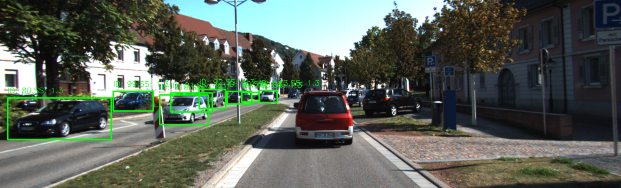}
    \end{minipage}%
    \begin{minipage}[b]{0.33\linewidth}
        \centering
        \includegraphics[width=\linewidth,trim={0cm 0.3cm 0cm 1.0cm},clip]{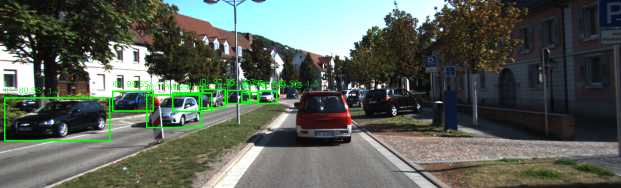}
    \end{minipage}\\
    \vspace{-0.3cm}
    
    \caption*{Query: Car on the left}
    \vspace{0.25cm}
    % Fourth row
    \begin{minipage}[b]{0.33\linewidth}
        \centering
        \includegraphics[width=\linewidth,trim={0cm 0.3cm 0cm 1.0cm},clip]{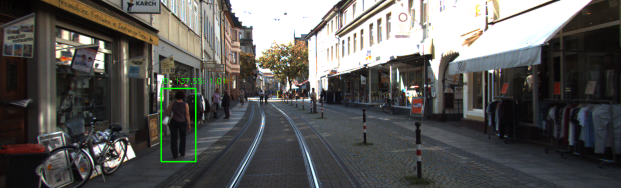}
    \end{minipage}%
    \begin{minipage}[b]{0.33\linewidth}
        \centering
        \includegraphics[width=\linewidth,trim={0cm 0.3cm 0cm 1.0cm},clip]{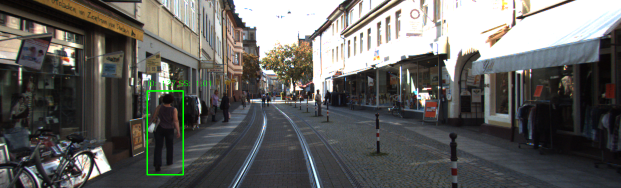}
    \end{minipage}%
    \begin{minipage}[b]{0.33\linewidth}
        \centering
        \includegraphics[width=\linewidth,trim={0cm 0.3cm 0cm 1.0cm},clip]{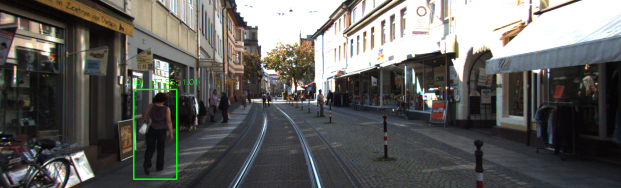}
    \end{minipage}\\
    \vspace{-0.3cm}
    \caption*{Query: A woman carrying a bag}

    % Main caption for all
    \caption{Qualitative results of our proposed ReferGPT on refer-KITTIv1~\cite{wu2023referring}. Each row presents example frames tracked according to the given text query.}
    \label{fig:qual_results}
\vspace{-0.5cm}
\end{figure*}

\subsection{Qualitative Results}
In figure \ref{fig:qual_results}, we illustrate qualitative examples of our proposed ReferGPT on the Refer-KITTIv1 dataset \cite{wu2023referring}. Each row presents frames from different video sequences where ReferGPT successfully tracks objects based on the input query. We accurately distinguish between similar objects in complex urban and highway scenes, maintaining robust tracking across frames, even under challenging occlusion cases.

\subsection{Ablation Study}

\noindent \textbf{Text Encoder Ablation.} We experiment with different distilled versions of text encoders to evaluate the trade-off between efficiency and performance in our matching module. As shown in Tab.\ref{tab:tab_encoders}, DistilBERT-L6 achieved a HOTA of 38.26\% while DistilBERT-L12 significantly improves the results, reaching 49.26\% HOTA, showing that increasing model depth enhances the embedding quality of the descriptions and queries. The CLIP encoder achieves the best performance of 49.46\% HOTA. These results suggest that distilled models can offer a promising balance between computational efficiency and accuracy.

\begin{table}[h!]
\caption{Ablation Study on distilled text encoders. DistilBert-L6 \cite{sanh2019distilbert} refers to a distilled version of Bert \cite{devlin2019bertpretrainingdeepbidirectional} with 6 layers, while Distil-L12 is the 12-layer model. The results are reported in \%.  \label{tab:tab_encoders}}
% \resizebox{\columnwidth}{!}{%
\centering
\begin{tabular}{c|ccc}
\toprule
\textbf{Encoder} & \textbf{HOTA}  & \textbf{DetA} & \textbf{AssA} \\
\midrule
DistilBert-L6 \cite{sanh2019distilbert} & 38.26 &  24.81 & 59.50 \\
DistilBert-L12 \cite{sanh2019distilbert} & 49.26 & 39.35 & 62.21 \\
CLIP \cite{radford2021learningtransferablevisualmodels} & \textbf{49.46} &  \textbf{39.43} & \textbf{62.57} \\
\bottomrule
\end{tabular}
\vspace{-0.5cm}
\end{table}

\noindent\textbf{Matching Module Ablation.} We compare different matching module configurations in Tab.~\ref{tab:tab_matching} to assess their contributions. The results show that using CLIP alone is insufficient achieving 31.13\% HOTA. Specifically, the CLIP Image encoder contributes the least, as it must encode a cropped detected object, which provides limited visual context. Additionally, the queries and descriptions often contain complex spatial and motion cues, making it difficult for the image encoder to establish strong similarities between the text and the image. Furthermore, since CLIP is trained on shorter image-caption pairs, the CLIP Text encoder struggles with long descriptions, as they contain words with high semantic weight such as the color, movement, or direction of the object. The best performance is achieved by combining CLIP Text with fuzzy matching, as this balances semantic understanding with token-level precision.

\begin{table}[h!]
\caption{Ablation Study on matching components. 'Clip Im' refers to the CLIP image encoder. 'Clip Text' denotes the CLIP text encoder, and 'Fuzzy' indicates the Fuzzy matching module. The results are reported in \%.\label{tab:tab_matching}}
\resizebox{\columnwidth}{!}{%
\centering
\begin{tabular}{ccc|ccc}
\toprule
\textbf{Clip Im} & \textbf{Clip Text} & \textbf{Fuzzy} & \textbf{HOTA}  & \textbf{DetA} & \textbf{AssA} \\
\midrule
\checkmark & - & - & 21.42 &  9.48 & 48.82 \\
- & \checkmark & -& 30.04 & 15.51 & 58.36 \\
\checkmark & \checkmark & - & 31.13 & 15.98 & 61.00 \\
- & - & \checkmark & 47.48 &  35.94 & 62.97 \\
\checkmark & - & \checkmark & 48.96 &  38.24 & 63.00  \\
\checkmark & \checkmark & \checkmark & 49.34 & \textbf{39.41} & 62.25  \\
- & \checkmark & \checkmark & \textbf{49.46} &  {39.43} & \textbf{62.57} \\
\bottomrule
\end{tabular}}
\end{table}

\noindent \textbf{Filtering Components Ablation.} We conduct experiments with different filtering strategies to evaluate their impact on performance. Applying a fixed similarity threshold, which renders our method online, achieves 36.61\% HOTA. Performing post-processing by clustering alone, increases our results to 46.5\%. By combining majority voting and clustering we achieve our best result.

\begin{table}[h!]
\caption{Ablation Study on  filtering components. \textbf{MV} refers to the majority voting process and \textbf{C} to the clustering. With \textbf{T}, we refer to a fixed threshold filtering. The results are reported in \%. \label{tab:tab_filtering}}
% \resizebox{\columnwidth}{!}{%
\centering
\begin{tabular}{ccc|ccc}
\toprule
\textbf{MV} & \textbf{C} & \textbf{T} & \textbf{HOTA}  & \textbf{DetA} & \textbf{AssA} \\
\midrule
 - & - & \checkmark & 36.61 & 23.17 & 58.23 \\
- & \checkmark & - & 46.50 & 36.92 & 59.10 \\
\checkmark & \checkmark & - & \textbf{49.46} &  \textbf{39.43} & \textbf{62.57}  \\
\bottomrule
\end{tabular}
\end{table}

\noindent \textbf{Object Detector Ablation.} We experiment with different object detectors, all pre-trained on KITTI, to evaluate their impact on RMOT performance. Our method is designed to be detector-agnostic and works effectively with both 2D and 3D detectors. While CasA achieves the highest 3D detection performance on KITTI among the tested detectors, leading to the best overall results in RMOT, the referring performance gap between CasA and lighter detectors such as QT-3DT (2D input) remains relatively narrow. This suggests that our referring text module is robust enough to compensate for lower-quality object detections. 
\begin{table}[h!]
\caption{Ablation Study on different 3D Object Detectors. The results are reported in \%.\label{tab:3d_detector}}
% \resizebox{\columnwidth}{!}{%
\centering
\begin{tabular}{cc|ccc}
\toprule
\textbf{Object Detector} & Input & \textbf{HOTA}  & \textbf{DetA} & \textbf{AssA} \\
\midrule
QT-3DT \cite{hu2022monocular} & 2D & 46.36  &   36.58   & 59.00  \\
PV-RCNN \cite{shi2020pv}  & 3D & 44.93 &  34.07 & 59.86 \\
Second-IOU \cite{yan2018second} & 3D & 45.01 &  33.97 & 60.14 \\
Point-RCNN \cite{shi2019pointrcnn}  & 3D & 47.14 &  36.04 & 61.92 \\
CasA~\cite{wu2022casa} & 3D & \textbf{49.46} &  \textbf{39.43} & \textbf{62.57} \\
\bottomrule
\end{tabular}
\end{table}

\noindent \textbf{Distilled MLLM Ablation.} 
Table~\ref{tab:3d_detector} shows the impact of different distilled MLLMs on RMOT performance. GPT-4o-mini achieves the highest HOTA, demonstrating its strong prompt-following and spatial reasoning capabilities. While smaller models like Qwen2 \cite{bai2023qwentechnicalreport} and Phi-4 \cite{abdin2024phi4technicalreport} perform worse in detection, their association accuracy remains relatively stable. However, the matching module alone does not suffice. For example, Qwen2 produces suboptimal detected object descriptions (Confusing object colors or pedestrians' genders), showcasing that the quality of the MLLM’s descriptive output remains essential for maximizing detection and overall tracking performance. This highlights the potential for even greater performance gains through fine-tuning.

\begin{table}[h!]
\caption{Ablation Study on different distilled MLLMs. The results are reported in \%.\label{tab:3d_detector}}
\resizebox{\columnwidth}{!}{%
\centering
\begin{tabular}{cc|ccc}
\toprule
\textbf{MLLM} & \textbf{Size}& \textbf{HOTA}  & \textbf{DetA} & \textbf{AssA} \\
\midrule
Qwen2-VL-Instruct \cite{bai2023qwentechnicalreport} & 2B & 19.97 & 6.50 & 61.66 \\
Phi4-Multimodal-Instruct \cite{abdin2024phi4technicalreport} & 5B & 36.56 &  21.49 & 62.56 \\
GPT-4o-mini$^{*}$ & 8B* & \textbf{49.46} &  \textbf{39.43} & \textbf{{62.57}}  \\

\bottomrule
% \begin{minipage}{\columnwidth}

% \end{minipage}
\end{tabular}}

\vspace{1mm}
\begin{minipage}{\columnwidth}
\footnotesize
$^{*}$ \textit{The exact model size has not been officially disclosed. The reported parameter count is based on publicly available information and third-party sources.}
\end{minipage}
\end{table}
\vspace{-0.5cm}
\subsection{Limitations}
While our method eliminates the need for retraining, the inference pipeline remains computationally expensive. The process of generating natural language descriptions for each detected object, computing similarity scores and performing query matching introduces additional latency. This computational overhead arises from the repeated use of the multimodal large language model, which is resource-intensive both in terms of processing time and memory consumption. However, our work can benefit from advancements in efficient or distilled MLLMs, which promise to reduce inference time and resource demands without sacrificing performance.

Another limitation is that, despite our results demonstrating a flexible vocabulary of referring expressions, our framework is still not fully open-vocabulary. Indeed, the MLLM-generated captions follow the structure and specificity of their prompting. The captions often contain information that may not always align with the level of abstraction required by a given query. As a result, if a user query refers to an aspect that is absent from the generated description, our method may struggle to establish a correct match. Achieving true open-vocabulary refer tracking requires exploring further zero-shot generalization, as it is infeasible to train a model on every possible user query. 

\section{Conclusion}
In this work, we present ReferGPT, a novel framework that performs zero-shot Referring Multi-Object Tracking in 3D. We overcome the limitations of current supervised approaches, that struggle with novel and ambiguous queries, showcasing the scalability and adaptability of our method. We leverage our 3D tracker and feed an MLLM 3D spatial information, enhancing its ability to generate structured, spatially aware descriptions within the tracking-by-detection paradigm. Our extensive evaluations on three different referring datasets based on KITTI traffic scenes, demonstrate that ReferGPT can generalize across diverse and open-set queries, highlighting its potential for tracking in complex, open-world scenarios.

\section*{Acknowledgement}
The work is supported by the ”Onderzoeksprogramma Artificiele Intelligentie (AI) Vlaanderen” programme and by Innoviris within the research project TORRES. N. Deligiannis acknowledges support from the Francqui Foundation (2024-2027 Francqui Research Professorship).

{
    \small
    \bibliographystyle{ieeenat_fullname}
    \bibliography{main}
}

% WARNING: do not forget to delete the supplementary pages from your submission 
% \input{sec/X_suppl}

\end{document}